\documentclass{article}
\usepackage{spconf,amsmath,graphicx, color, soul, multirow}

\usepackage{amssymb}
\usepackage{subfig}
\usepackage{bm}
\usepackage{verbatim}
\usepackage{cite}
\usepackage{amssymb}
\usepackage{pifont}
\newcommand{\cmark}{\ding{51}}%
\title{Towards Language-Universal End-to-End Speech Recognition}
\twoauthors
  {Suyoun Kim\sthanks{The author performed the work during an internship at Microsoft Research.}}
{Dept. of Electrical \& Computer Engineering \\
Carnegie Mellon University \\
Pittsburgh, PA \\
\texttt{suyoun@cmu.edu} }
 {Michael L. Seltzer}
	{Speech \& Dialog Research Group \\
   	Microsoft AI \& Research\\
	Redmond, WA 98052 \\
    \texttt{mseltzer@microsoft.com}}

\begin{document}
\ninept
\maketitle
\begin{abstract}

Building speech recognizers in multiple languages typically involves replicating a monolingual training recipe for each language, or utilizing a multi-task learning approach where models for different languages have separate output labels but share some internal parameters. In this work, we exploit recent progress in end-to-end speech recognition to create a single multilingual speech recognition system capable of recognizing any of the languages seen in training. To do so, we propose the use of a universal character set that is shared among all languages. We also create a language-specific gating mechanism within the network that can modulate the network's internal representations in a language-specific way. We evaluate our proposed approach on the Microsoft Cortana task across three languages and show that our system outperforms both the individual monolingual systems and systems built with a multi-task learning approach. We also show that this model can be used to initialize a monolingual speech recognizer, and can be used to create a bilingual model for use in code-switching scenarios.

\end{abstract}
\section{Introduction}
\label{sec:intro}
As voice-driven interfaces to devices and information become mainstream, increasing the global reach of speech recognition systems becomes increasingly important. There are two primary challenges that arise in expanding the language coverage of a speech application. First, since conventional speech recognition systems require each model to be trained independently, as the number of supported languages grows, the effort required to train, deploy, and maintain so many models in a production environment will increase dramatically. In addition, for second- and third-tier languages with fewer resources available, issues with data scarcity arise. For each language, building a speech recognition system requires a large collection of transcribed speech recordings from many speakers to train an acoustic model, linguistic expertise to create a pronunciation dictionary, and vast amounts of text data to train a language model. 


Over the years, prior work has attempted to address the issues faced by low-resource language via a transfer learning approach \cite{lin2009study, tuske2013investigation, ghoshal2013multilingual, huang2013cross, heigold2013multilingual, miao2014improving, besacier2014automatic, gales2015unicode}.  In these approaches, language-specific deep networks are trained in which the the parameters in the lower layers of the network are shared across languages. This approach can also be interpreted as an instance of multi-task learning, where information across tasks, i.e. languages, is shared to create a more informative internal representation, less prone to over-fitting. Another common approach for creating models in low resource languages is to adapt a neural acoustic model that has been well trained on a high-resource language. This is typically done by replacing output layer of the well-trained model and re-training the model to predict the targets of low-resource languages \cite{tuske2013investigation, ghoshal2013multilingual, huang2013cross, vu2013multilingual, heigold2013multilingual, wang2015transfer}. 
All of these models have been based on the conventional acoustic modeling  strategy based on senones, and therefore still require a pronunciation lexicon to map words to phonemes and then senones.

Recently, new approaches to speech recognition that work in a so-called \emph{end-to-end} manner have been proposed. In these systems, a neural network is trained to convert a sequence of acoustic feature vectors into a sequence of graphemes rather than senones. Unlike sequences of senone predictions, which need to be decoded using a pronunciation lexicon and a language model, the grapheme sequences can be directly converted to word sequences without any additional models or machinery. The end-to-end models proposed in the literature operate using a Connectionist Temporal Classification framework \cite{graves2006connectionist, graves2014towards, hannun2014deep, miao2015eesen, zweig2017advances}, an attention-based encoder-decoder framework \cite{bahdanau2014neural, chorowski2014end, chorowski2015attention, chan2015listen}, or both \cite{kim2017joint}. 
Thus far, the research in end-to-end systems has focused on monolingual scenarios and to the best of our knowledge, there have been no studies of grapheme-based models capable of recognizing multiple languages. There has been recent work that shows how a low-resource graphemic system can be initialized with a well-trained high-resource model, 		as was done for the senone-based models described previously \cite{kunze2017transfer}. 


In this paper, we make significant progress towards a language-universal speech recognizer by creating a single end-to-end system capable of recognizing any language it has been trained on. Our model exploits the recent progress in end-to-end approaches to output character sequences directly, without requiring pronunciation lexicons. As in other multilingual systems, we apply a transfer learning approach to share model parameters among multiple languages. The novel aspects of our proposed model are twofold. First, we use a single universal character set that can be shared among all languages  rather than separate language-specific output layers. Second, we propose a language-specific gating mechanism in the network that can increase the network's modeling power by using multiplicative interactions to modulate the network's internal representations in a language-specific way. We evaluate our proposed model on the Microsoft Cortana personal assistant task and show that our system outperforms separate language-specific models as well as the conventional multi-task learning approach. 

The remainder of this paper is organized as follows. In Section \ref{sec:ctc}, we briefly review the end-to-end speech recognition paradigm. In Section \ref{sec:multilingual_ctc}, we describe approaches to multilingual  CTC including multi-task learning and our proposed model that features end-to-end learning with a universal character set and language-dependent gating units. In Section \ref{sec:exp}, we evaluate our proposed model through a series of experiments on the Microsoft Cortana task across three languages. Finally, we summarize our findings and discuss future avenues of research in Section \ref{sec:conclusion}.

\section{End-to-end modeling using CTC}
\label{sec:ctc}

\begin{figure}[t]
\centering

\begin{minipage}[b]{1.0\linewidth}
\subfloat[Separate labels via Multi-task Learning]{\includegraphics[width = 1.5in, height = 1.6in]{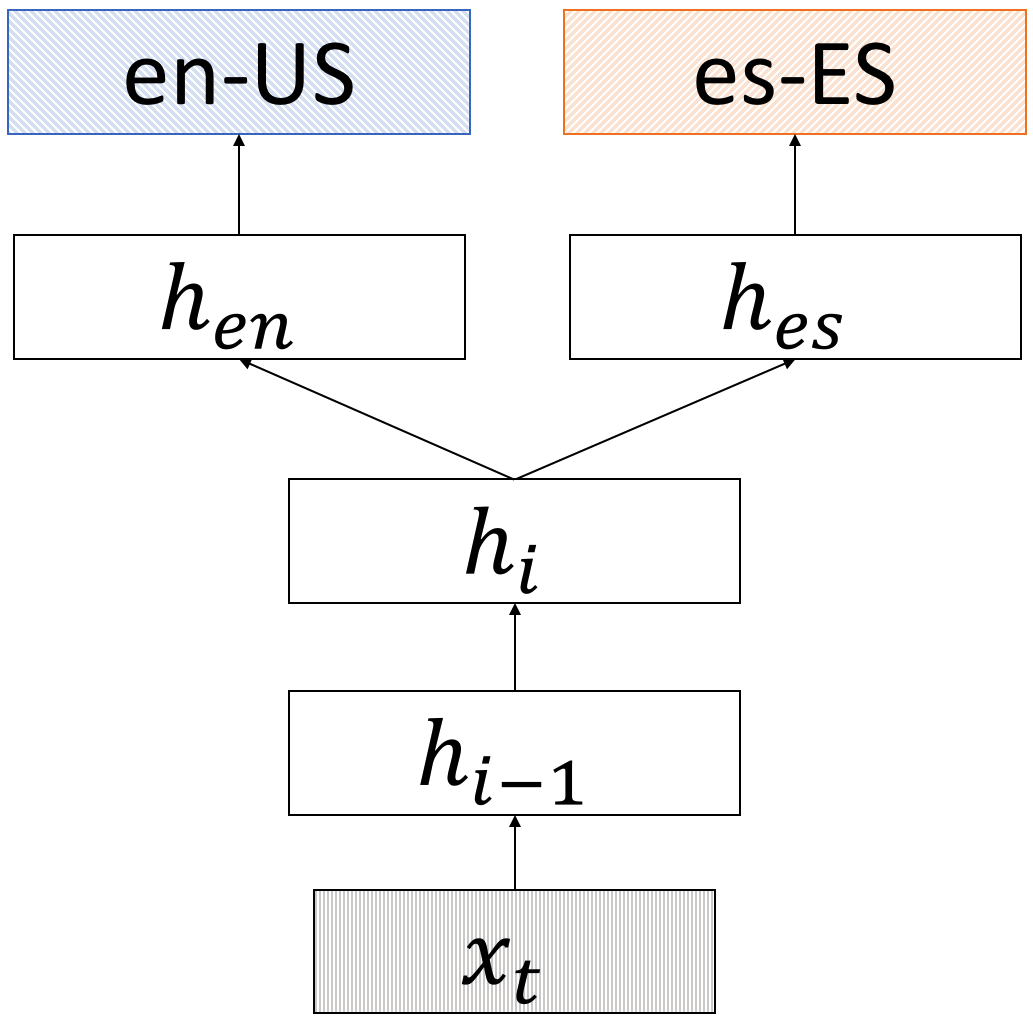}} 
\hspace{10pt}
\subfloat[Universal label set with masking]{\includegraphics[width = 1.5in, height = 1.6in]{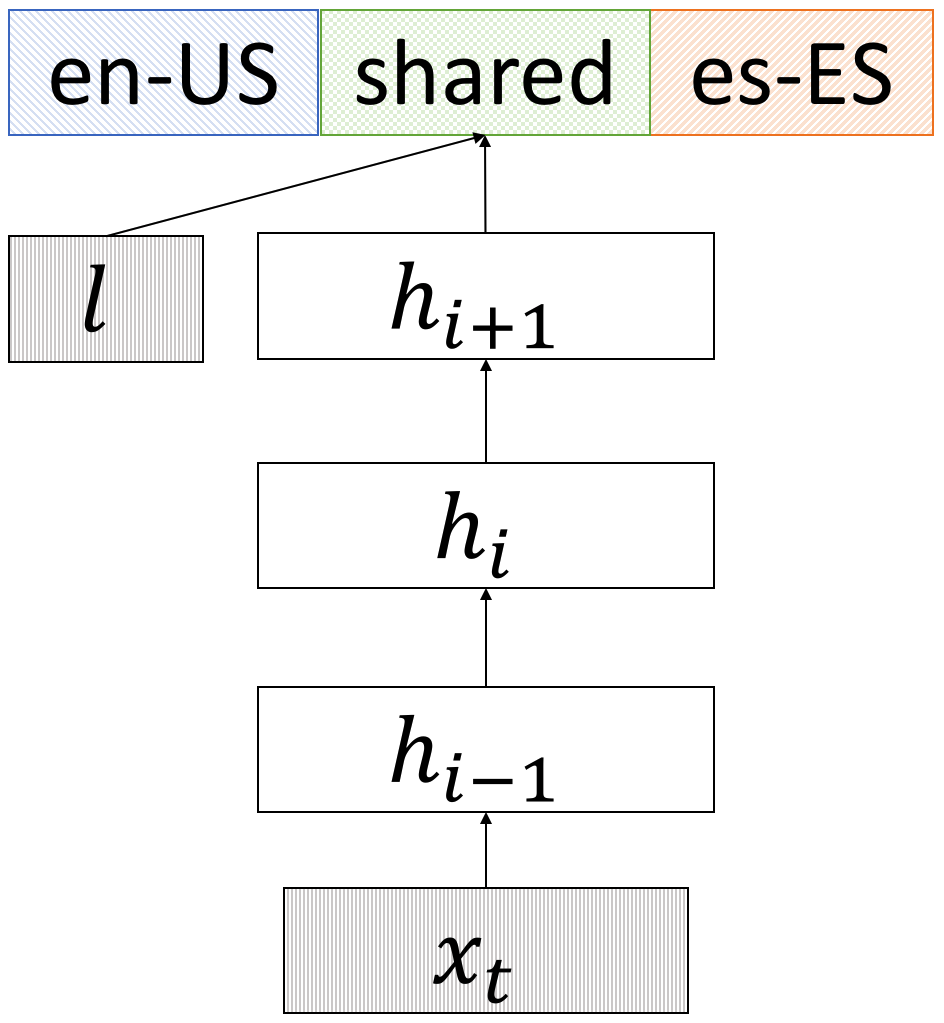}} 
\end{minipage}

\caption{The two different architectures of language-universal end-to-end speech recognition models (a), (b) and our proposed language-specific gating (c).} 
\label{fig:arch}
\end{figure}

In this work, we perform end-to-end speech recognition using a CTC-based approach with graphemes (characters) as the output symbols. With CTC, the neural network is trained according to a maximum-likelihood training criterion computed over all possible segmentations of the utterance's sequence of feature vectors to its sequence of labels \cite{graves2006connectionist}. Core to CTC training is the presence of the \emph{blank} symbol which can be interpreted as a ``don't care" symbol in the output label sequence. In CTC, the label sequences can contain blanks and repeated characters without penalty.

Given a sequence of acoustic feature vectors, $\bm x$ and the corresponding graphemic label sequence, $\bm y$, CTC trains the model to maximize the probability distribution over all possible label sequences $\bm{\pi}$. 
    \begin{align}
	P(\bm{y}|\bm{x}) = & \sum_{\bm{\pi} \in \Phi(\bm{y})}P(\bm{\pi}|\bm{x}) \approx \sum_{\bm{\pi} \in \Phi(\bm{y})} \prod_{t=1}^T P(\pi _t|\bm{x}).
    \end{align} 
Likelihoods are computed using the well-known forward-backward algorithm and the gradient of the likelihood is used to updated the network parameters using back propagation. After model training, decoding is performed in a greedy manner, where the most likely symbol at each time frame is hypothesized. The final sequence is obtained via a post-processing step where any character repetitions and blank symbols are removed from the output. Further improvements can be obtained by incorporating an external language model, at the character or word level \cite{graves2014towards, hannun2014deep, miao2015eesen, zweig2017advances}.

In this work, we consider two neural architectures for a language-universal end-to-end model that both use the CTC objective function.

\section{Multilingual end-to-end models}
\label{sec:multilingual_ctc}
\subsection{Multi-task CTC with language-specific character sets}


Multi-task learning (MTL) has been proposed as a means for improving generalization performance in low-data scenarios. Training a model on multiple related tasks simultaneously serves as an inductive bias to improve the model's performance. Many prior approaches to training acoustic models for low resource languages have used MTL as a means of sharing some of the acoustic model parameters across languages, e.g. all the parameters up to the final output layer \cite{lin2009study, heigold2013multilingual, miao2014improving}. This forces the model to learn commonalities across languages which provides effective regularization and prevents over-fitting.  To the best of our knowledge, this is the first effort to evaluate MTL with an end-to-end model and a CTC objective function. Our MTL architecture is similar to previous senone-based models, where the lower layers are shared among the different languages while the output layer or layers are trained to be language specific. This architecture is shown in Figure \ref{fig:arch}a. Note that while the figure shows a language-specific output layer and final hidden layer, the choice of where to branch the model is a design choice that should be determined by experimentation. 

Multi-task learning for CTC can be realized as a combination of the individual single-task objective functions. We define $l$ to be an index over the different languages in the training data, $\bm{x}_{l}$ as a sequence of input feature vectors, and $\bm{y}_l$ the corresponding grapheme sequence. We can then define the CTC objective function for the $l$th language as 
    \begin{equation}
	\mathcal{F}_{l}(\theta_s,\theta_l) \triangleq -\ln P(\bm{y}_{l}|\bm{x}_{l},\theta_s,\theta_l)
    \end{equation} 
where $\theta_l$ are the language-specific model parameters, and $\theta_s$ are the model parameters shared across languages. The multi-task objective function over all languages can then be defined as the summation of the individual negative log likelihoods of all $L$ languages,
    \begin{equation}
	\mathcal{F} \triangleq \sum_{l=1}^{L} \mathcal{F}_{l}(\theta_s,\theta_l)
    \end{equation} 
%


\subsection{Multilingual CTC with a universal character set}
\label{sec:union}

In many instances, present-day languages evolved from a common ancestry. It is therefore natural that they share some common graphemes and phonemes. For example, the English character set is a subset of the Spanish character set and knowing one language helps to speak and write the other. With this as motivation, we propose a multilingual architecture that uses a ``universal" output label set consisting of the union of all characters from the multiple languages, as illustrated in Figure \ref{fig:arch}b. Unlike the MTL approach, this is a single model with a universal character set that is trained on all multilingual data. Characters that are common across multiple languages are trained based on all relevant data while language-specific characters are only trained with data from that language. 

Given the training data from multiple languages and the universal label set, $\bm{y}_{u}$, the CTC loss for the models to be minimized is defined as,
    \begin{equation}
	\mathcal{L}_{U}(\theta) \triangleq -\ln P(\bm{y}_{u}|\bm{x}_1,\dots,\bm{x}_L, \theta)
    \end{equation} 
%

In this model, we assume \emph{a priori} that we know the language identity of the utterances in both training and decoding. We use this information to restrict the predictions to only those characters present in the corresponding language and mask the activations from the other irrelevant characters.
Let $Y_u$ be the universal label set with $K$ distinct labels. The label set for any particular language $Y_l$ is a subset of $Y_u$ ($Y_l \subseteq Y_u$). 
Given a language indicator, $l$, we can mask out the activations from unwanted characters using a $K$-dimensional binary mask defined as  
%
    \begin{align}
    M[ l, k ] =&  
            \begin{cases}
            0, \text{if} \; k \not\in Y_{l} \\
            1, \text{if} \; k \in Y_{l}
            \end{cases}
    \end{align}
This mask is then applied to the network outputs to compute the log-likelihoods for the CTC objective function. Note that this masking operation is applied in both training and decoding. 

\subsection{Language-specific gating units}
\label{sec:arch2}
One possible drawback to the proposed universal character approach is that the same grapheme may have different underlying phonetic realizations in different languages. For example, the letter $i$ in English typically corresponds to the sound /ih/ in English, but /iy/ in Italian. Thus, the model needs to adequately capture language-specific information in order to properly account for such differences. Adding a language identification feature as an auxiliary input to the model is the simplest way to do so. However, we found empirically that this only provides minimal improvement. Instead, we propose to use the language identity to modulate the network's internal multilingual representations in a language-specific manner. To do so, the outputs of each hidden layer are processed by a series of language-dependent gates before being passed to the next layer in the model. Specifically, we first create a one-hot language indicator vector $d_l$ for each language $l$,


%
    \begin{equation}
    \label{e1}
	d_l = [0 \hspace{0.2cm} 0 \hspace{0.2cm} 1].
    \end{equation}
%
Then, we compute the gate value based on the language indicator vector $d_l$ and the current output values of $h_i$, the $i$th hidden layer
	\begin{equation}
    \label{e2}
    g(h_i, d_l) = \sigma(U h_i + V d_l + b)	
	\end{equation}
where $U, V$, and $b$ are trainable parameters. 
The language-gated hidden activations are then calculated as
	\begin{equation}
    \label{e3}
    \hat {h_i} = g(h_i, d_l) \odot h_i
	\end{equation}
Finally, $\hat {h_i}$ and $d_l$ are concatenated and input to the next layer. 
	\begin{equation}
    \label{e4}
    \tilde {h_i} = [\hat {h_i} : d_l]
	\end{equation}

\section{EXPERIMENTS}
\label{sec:exp}

\subsection{Experimental corpora}

We investigated the performance of the proposed language-universal model on English (EN), German (DE), and Spanish (ES) data from Cortana, Microsoft's personal assistant. For each language, we used 150 hours of training data, 10 hours of validation data, and 10 hours of test data. We used 80-dimensional log-mel filterbank coefficients as acoustic features, derived from 25~ms frames with a 10~ms frame shift. We concatenate three consecutive feature vectors to input to the network and employ frame-skipping \cite{sak2015fast} which decimates the original frame rate by a factor of three. Thus, each feature vector is presented to the network exactly once. Following \cite{zweig2017advances}, we used a label symbol inventory consisting of the individual characters and their double-letter units. An initial capitalized letter rather than a space symbol was used to indicate word boundaries. This resulted in 81 distinct labels for English, 93 labels for German, and 97 labels for Spanish. Our universal label set for these three languages had 108 distinct labels and 81 overlapping labels. No pronunciation lexicon or language model was used in any of the experiments.    

\subsection{Training and decoding}

Our language-universal encoder was a 4-layer Bidirectional Long Short-Term Memory (BLSTM) network \cite{hochreiter1997long, graves2013hybrid} with 320 cells in each layer and direction. A linear projection layer followed each BLSTM layer. All the weights in the models were initialized with a uniform distribution in the range of [-0.05, 0.05], and were trained using stochastic gradient descent with momentum. We used a learning rate of 0.0004 and gradient-clipping threshold per sample of 0.0003. Early stopping on the validation set was used to select the best model. For the decoding, the most likely sequence of characters was generated by the model in a greedy manner. The final output sequence was then obtained by removing any blank symbols or repetitions of characters from the output and replacing any capital letter with a space and its lowercase counterpart. 

\subsection{Gated language-universal end-to-end models}
\begin{table}[t]
\caption{ Initial CER results for conventional single-task (stl), multi-task (mtl), and universal character set (univ) networks using monolingual and multilingual training data. }
  \vspace{-0.5cm}
\label{tab:result1}
\begin{center}
\begin{tabular}{l|c|c| c| c c }
  \hline
 Training & Total & Model & Test &  CER & \% Rel. \\
Languages & Hrs  & Arch & Lang & \%         & Impr.        \\  
\hline \hline
DE & 150 & stl & \multirow{5}{*}{DE} & 23.3 & -   \\
DE + EN & 300 & mtl & & 22.3 & 4.0 \\
DE + EN & 300 & univ & & 22.5 & 3.2 \\
DE + EN + ES & 450 & univ & &22.8 & 2.1 \\
DE & 300 & stl & &\textbf{15.8} & \textbf{32.2} \\ 
\hline 
ES & 150 & st				& \multirow{5}{*}{ES} & 13.7 & -   \\
ES + EN & 300 & mtl 		& & 13.1 & 4.4 \\
ES + EN & 300 & univ 	& & 12.9 & 5.8 \\
ES + EN + DE & 450 & univ 	& & 13.1 & 3.9 \\
ES &300 & stl				& & \textbf{11.7} & \textbf{14.4} \\
\hline
\end{tabular}
\end{center}
\end{table}

We first evaluated both multi-task and language-universal architectures on German and Spanish data, along with conventional monolingual training. As seen in Table \ref{tab:result1}, we obtained a small performance gain over the monolingual model by using multilingual training data. In addition, we observed that models trained with multi-task learning with language-specific output labels and models trained with a universal label set performed comparably, demonstrating the potential of a universal label set. However, with a conventional BLSTM topology, the benefit gained from the multilingual data was far less than simply doubling the amount of training data from the target language to 300 hours. More concerning, the performance actually degraded when we increased the multilingual data from two to three language. 
We also extensively evaluated various configurations for multi-task learning, where all or only some of the layers were shared, and found no significant improvements over the results reported in Table \ref{tab:result1}. These initial results indicate that simply training conventional models with multilingual data is not very helpful. However, we note that these results may differ in extremely low resource scenarios where the data available per language could be an order of magnitude smaller.  


\begin{table}[t]
\caption{Performance of the language-universal modeling approach with and without the language-specific gating. Baseline monolingual performance is also shown. Each language has 150~h of training data.}
  \vspace{-0.5cm}
\label{tab:result2}
\begin{center}
\resizebox{\linewidth}{!}{
\begin{tabular}{ c | c | c c c | c c c }
  \hline
Training & Lang & \multicolumn{3}{c}{\% CER} & \multicolumn{3}{|c}{\% WER} \\
Data     &  	Gate &EN & DE & ES & EN & DE & ES \\  \hline \hline
monolingual & --  & 20.3 & 23.3 & 13.7 & 45.0 & 57.3 & 39.9 \\
\hline
\multirow{2}{*}{EN + DE} &  --  	      & 19.7 & 22.5 & -- &45.1 & 56.3 &	--\\
  &   \cmark   & 18.4 & 21.0 & -- & 42.2 & 53.4 &--	\\
 \hline
\multirow{2}{*}{EN + ES} & -- 		  & 19.9 & --    & 12.9 & 45.5 & -- & 38.7	\\
 &   \cmark & 18.4 & --    & 12.1 & 42.2 & -- & 35.9\\
\hline
\multirow{2}{*}{EN + DE + ES} &  --     & 19.9 & 22.8 & 13.1 & 46.3 & 57.6 &	40.5\\
 & \cmark & \textbf{18.1} & \textbf{20.6} & \textbf{11.7} & \textbf{41.9} & \textbf{52.4} & \textbf{35.5}\\ 
\hline
\end{tabular}
}
\end{center}
\end{table}
We next evaluated the proposed language-universal model with language-specific gating. The results in Table \ref{tab:result2} show that the proposed gated model significantly outperforms the monolingual model as well as the conventional multilingual models. Our model showed 10.7\%, 11.4\%, and 14.1\% relative improvement in CER compared to a monolingual for EN, DE, and ES respectively. The relative improvements in WER, which was calculated without using any language model, were 7.0\%, 8.6\%, and 11.1\%, for EN, DE, and ES respectively. Furthermore, when language-specific gating is used, additional improvement is obtained when the number of training languages increased from two to three. 

\begin{figure}[!h]
\begin{minipage}[b]{1.0\linewidth}
  \centering
  \centerline{\includegraphics[width=8.5cm]{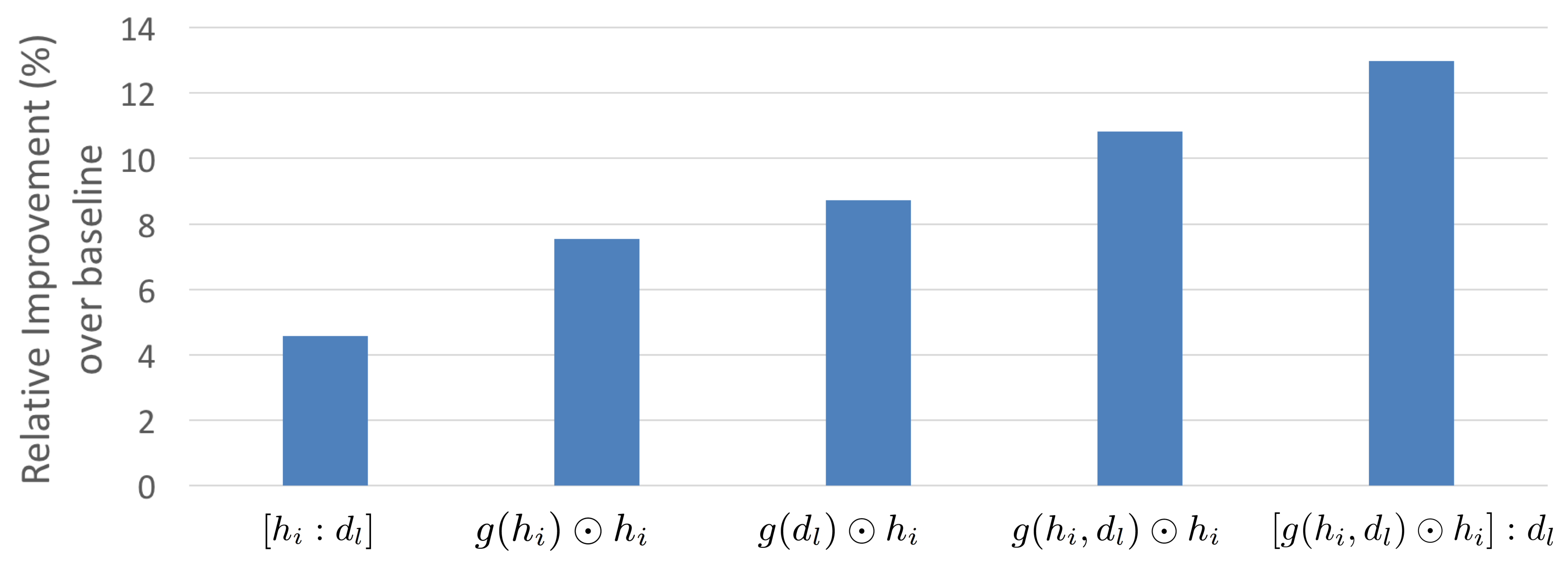}}
\end{minipage}
\caption{The relative improvement in WER obtained by using language identification information as an auxiliary input and/or a gating mechanism. 
}
\label{fig:gate_function_analysis}
\end{figure}

Figure \ref{fig:gate_function_analysis} shows the relative improvement over the baseline WER obtained by adding language information to the model in various ways. The leftmost bar shows the result of simply augmenting the input to each layer with the one-hot language vector $d_l$. The next three bars show the performance of different gating functions driven by the current hidden state, the language identity, or both, respectively. Finally, the rightmost bar shows the approach shown in Equations \ref{e3}-\ref{e4}. As the figure indicates, all approaches provide gains over the baseline model, but the proposed approach results in the largest improvement. We also investigated the effect of the gating mechanism on different layers and observed that the best performance was obtained when gating is applied to every layer (not shown). 


\subsection{Initial model for subsequent language-specific models}
While the primary goal of this work was to create a language-universal model, we found that this model is also a good initial model for creating a language-specific monolingual model when the training data is limited. Table \ref{tab:result3} shows the CER on the DE test set obtained from different pre-training strategies used to initialize the model prior to fine-tuning on the DE training data. Notably, initializing with the gated language-universal model outperformed all of the other approaches, including bootstrapping from an EN model trained with significantly more data. The second best performance was obtained by the gated language-universal model directly, even without further fine-tuning on the DE training data. 

\begin{table}[t]
\caption{CER obtained for a monolingual model with different pre-training strategies. The model was then fine-tuned on 150 hours of the target language (DE), except * where no fine-tuning was performed.}
  \vspace{-0.5cm}
\label{tab:result3}
\begin{center}
\begin{tabular}{ c |c|| c }
  \hline
\multicolumn{2}{c||}{Pre-training} & \% CER             \\ \cline{1-2}
Training Data & Hours       &          DE    \\ \hline \hline
--             &       --        & 23.3                        \\
EN & 150                        & 24.3                        \\
EN &1000                        & 21.4                        \\
EN + DE & 300                	& 21.1                        \\
EN + ES + DE + gate & 450	& \textbf{19.4}~/~\textbf{20.6}* \\ 
\hline
\end{tabular}
\end{center}
\end{table}


\subsection{Bilingual end-to-end models}

One promising aspect of a universal output character set is the potential to create end-to-end systems that can dynamically code-switch between languages. This can be done by training a model on bilingual training data with the union of the output symbols of the two languages while omitting the language-specific aspects of the model. 
Following this approach, we created a bilingual model from 150 hours of English (EN) and 150 hours of Spanish (ES) using the union of the character sets from the two languages. 

Because a test set of bilingual mixed-code speech was not available, we evaluated this model on the EN and ES test sets separately. Table \ref{tab:result4} shows that bilingual models perform as well as the monolingual models even if no language identification information is provided to the network. Thus, this approach is very promising as a method for creating end-to-end systems capable of code-switching during decoding. 

\begin{table}[h]
\caption{ WER on ES and EN test sets obtained from monolingual and bilingual end-to-end systems.}
  \vspace{-0.5cm}
\label{tab:result4}
\begin{center}
\begin{tabular}{ c| c | c c }
\hline
\multirow{2}{*}{Training Data} & \multirow{2}{*}{Hours} & \multicolumn{2}{c}{\% WER} \\ 
 & &  ES & EN \\ \hline
monolingual EN or ES & 150	& 39.9 	& 45.0 \\
bilingual EN + ES w/o Gate & 300 & 38.8 	& 45.1 \\
\hline
\end{tabular}
\end{center}
\end{table}

\section{CONCLUSION}
\label{sec:conclusion}

We proposed a language-universal end-to-end speech recognition system capable of recognizing speech from any language seen in training. Key aspects of this model include the use of a universal character set and language-specific gating units. Because this model can support multiple languages, it has the potential to simplify model deployment in production environments. In addition, no pronunciation lexicon is required, which is beneficial for second- and third-tier languages, where such resources may be unavailable or limited. Our model was shown to outperform models trained on monolingual training data, as well as with the multi-task learning approach employed in previous work. Moreover, we found that our language-universal speech recognizer is a strong initial model for subsequent monolingual training, and demonstrated the potential of this system for decoding speech in the presence of code-switching. Moving forward, we plan to explore how to add a new language to an already-trained multilingual system and evaluate the performance improvements that can be obtained by using an external language model. 



\vfill\pagebreak

\bibliographystyle{IEEEbib}
\bibliography{strings,refs}

\end{document}